\documentclass[graybox]{svmult}

\usepackage{mathptmx}       
\usepackage{helvet}         
\usepackage{courier}        
\usepackage{type1cm}        
\usepackage{makeidx}         
\usepackage{graphicx}        
\usepackage{multicol}        
\usepackage[bottom]{footmisc}

\makeindex             

\begin{document}

\title*{Formal Concept Analysis for Knowledge Discovery from Biological Data}
\author{Khalid Raza}
\institute{Khalid Raza \at Department of Computer Science, Jamia Millia Islamia (Central University), New Delhi, India  \email{kraza@jmi.ac.in}}
%
%
\maketitle

\abstract*{Due to rapid advancement in high-throughput techniques, such as microarrays and next generation sequencing technologies, biological data are increasing exponentially. The current challenge in computational biology and bioinformatics research is how to analyze these huge raw biological data to extract biologically meaningful knowledge. This chapter focuses on the applications of formal concept analysis for the analysis and knowledge discovery from biological data, including gene expression discretization, gene co-expression mining, gene expression clustering, finding genes in gene regulatory networks, enzyme/protein classifications, binding site classifications, and so on. It also presents a list of FCA-based software tools applied in biological domain and covers the challenges faced so far.}

\abstract{Due to rapid advancement in high-throughput techniques, such as microarrays and next generation sequencing technologies, biological data are increasing exponentially. The current challenge in computational biology and bioinformatics research is how to analyze these huge raw biological data to extract biologically meaningful knowledge. This chapter presents the applications of formal concept analysis for the analysis and knowledge discovery from biological data, including gene expression discretization, gene co-expression mining, gene expression clustering, finding genes in gene regulatory networks, enzyme/protein classifications, binding site classifications, and so on. It also presents a list of FCA-based software tools applied in biological domain and covers the challenges faced so far.}

\section{Introduction}
\label{sec:1}
After Human Genome Project, there is unprecedented growth in biological data. Due to technological advancement in high throughput technologies,
 such as Microarray and Next Generation Sequencing, it is possible to produce high quality biological data with rapid speed. The biological
 data can be broadly classified as Genomics, Transcriptomic and Proteomics. For example, gene expressions are transcriptomic data that quantify
 the state of genes in a cell. When these gene expression data are analyzed properly, it may reveal many hidden cellular processes and
 biological knowledge. These knowledge discoveries from biological data may lead to better understanding of disease mechanism and further it
 guide for better diagnosis and therapy of the disease.

Formal Concept Analysis (FCA), introduced by R. Wille in early 1980s \cite{wille82}, is a method based on lattice theory for the analysis
of binary relational data. Since its inception, FCA has been found to have potential applications in many areas including data mining, knowledge
discovery and machine learning. Like other computational technique, FCA has also been applied in microarray analysis, gene expression mining,
gene expression clustering, finding genes in gene regulatory networks, enzyme/protein classifications, binding site classifications, and so on.
In this chapter, we will present the current status
of FCA for the analysis and knowledge discovery from biological data and also cover challenges faced so far.

\section{Biological Databases}
\label{sec:2}
Due to availability of high-throughput techniques, biological database are being generated exponentially and
 the modern biology has turned into a data-rich science. Some of the important biological data are nucleotide
 and protein sequences, protein 3D structure produced by X-ray crystallography and NMR, metabolic pathways,
 complete genomes and maps, gene expression and protein-protein interaction, and so on.

Biological databases are broadly divided into sequences databases and structure databases.
Sequence data are applicable for both DNA and protein, but structural databases are applicable
for proteins only. Today, most of the biological databases are freely available to the researchers.
In general, biological databases can be classified as primary, secondary and composite databases.
A primary databases stores information of either sequence or structure. For example, Uni-PROT and PIR for protein sequence,
GenBank and DDBJ for Genome sequence and the Protein Databank for protein structure. Secondary database stores information which
are derived from the primary database source, such as conserved sequence information, active site residues of the protein families
arrived by multiple sequence alignment of a set of related proteins, etc. The SCOP, CATCH, PROSITE are few examples of secondary databases.
A composite database is a collection of variety of different primary database sources that avoid the need for searching into multiple database
 sources. The National Centre for Biotechnology Information (NCBI) is the main central host that links multiple database sources and makes
  these resources freely available to us. For more information about biological databases,
  refer tutorials in \cite{babu97} and \cite{kraulis01}.

\section{Microarray Analysis}
\label{sec:3}
\subsection{Mining gene expression data}
Due to rapid advancement in high throughput technology such as Microarray and Next Generation Sequencing, transcriptomic data has been produced
 in unprecedented way. But analysis and interpretation of these data remains a challenge for the researchers due to complexity of the biological
  systems. The motivation and biological background that need to be considered for gene expression mining are: i) mostly single gene participates
  in many biological processes, i.e., it has several functions, ii) a biological process implies a small subset of genes, iii) a biological process
  of interest may be active in many, all or none situation for a given dataset, and iv) differentially expressed genes over different samples are not frequent.

In transcriptomic, researchers routinely analyze expression level of genes in different situations such as in tumor samples versus normal samples.
 Formal Concept Analysis (FCA) has been successfully applied in the field of transciptomics. Some of the studies identified set of genes that are
 sharing same transcriptional behavior using FCA \cite{rioult03, rioult03a, kaytoue-uberall09}. Due to availability of
  large gene expression datasets, it is possible to apply data mining tools to identify patterns of interest in the gene expression data. One of the
   most widely used data mining technique is \emph{association rules} which can be applied for the analysis of gene expression data. Association rules may uncover
    biologically relevant associations between genes, or between different environmental conditions and gene expression. An association rules can be written
    in the form $S_1$ $\rightarrow$ $S_2$, where $S_1$ and $S_2$ are disjoint sets of data items. The set $S_2$ is likely to occur whenever the set $S_1$ occurs. Here, the data items may
    include highly expressed or repressed genes, or other relevant facts stating the cellular environment of genes such as diagnosis of a disease samples.
     Association rules mining has been applied for gene expression data mining by many researchers including \cite{creighton03}.

In this section, we have discussed only the application of different variants of FCA for gene expression data mining, especially extracting co-expressed groups of
genes sharing similar expression. Most of the methods for co-expressed genes mining are based on binary biclustering methods. Here, scaling of data is done using
 a single threshold and one expression value. The expression values above this threshold are considered as over-expressed and represented by 1; otherwise it is
 considered as under-expressed and represented by 0. Once, the gene expression values are discretized to binary table then strong relationships can be extracted
having biologically meaningful information. Kaytoue-Uberall et al. (2008) \cite{kaytoue-uberall08} proposed interval-based FCA to extract groups of co-expressed genes. Given a set of genes $G$,
a set of relationships $S$, and set of ordered intervals $T, (g, (s, t)) \in I$, where $g \in G, s \in S, t \in T$ and $I$ is binary relation means gene expression value of gene
 $g$ is interval of index $t$ for situation $S$. Hence, formal concept of the context $(G, S \times T, I)$ shows groups of genes having $G \in V$ in same interval. Although a priori
   determination of these intervals are difficult.

Messai et al. (2008) \cite{messai08} proposed interval-free FCA based method to cluster gene expression values. However, this algorithm does not deal with large data set and also no link to
interordinal scaling was done. To overcome these problems, Kaytoue-Uberall et al. (2009) \cite{kaytoue-uberall09} introduced two FCA-based methods for clustering gene expression data. The first method is
based on interordinal scaling and second one is based on pattern structures that require adaptation of algorithm computed with interval algebra. Between these two algorithms
by Kaytoue-Uberall et al. (2009) \cite{kaytoue-uberall09}, second method has been proved to be more computationally efficient and provide more readable results.
These algorithms have been tested on
 microarray gene expression data of fungus Laccaria biocolor taken from Gene Expression Omnibus databases (GSE9784) composed of 22,294 genes and five different conditions.
 For the dimension reduction, cyber-T \cite{kayala12} tool was used that filter dataset and returned 10,225 genes.

DNA methylate affects the expression of genes and their regulation may cause several cancer-specific diseases. It is observed in many investigations that
hypomethylation of DNA have been associated with many cancers including breast cancer. Amin et al., (2012) \cite{amin12} applied FCA for mining the hypomethylated genes among breast
cancer tumors. They constructed formal concepts lattices with significant hypomethylated genes for every breast cancer subtypes. The constructed lattice reflects the
biological relationships among breast cancer tumor subtypes. The proposed filter method has two stages: non-specific filter and specific filters. The non-specific filtering
 step determines the hypomethlated CPGs by computing the difference between the mean of methylation level for the corresponding adjacent normal tissue. The second stage
 (specific filtering) receives the output of the first stage as input and performs one-sample Kolmogorov Smirnov test to check the normality of each breast cancer subtypes.
 If the given dataset follows normal distribution then paired t-test is applied, otherwise Wilcoxon signed ranked is applied. Once, the filtering of hypomethylated genes
 is done then FCA has been applied to determine breast cancer subtypes. Here, Java-based FCA analysis software tool, called ConExp \cite{serhiy00} was used to generate the
  lattice diagram.

\subsection{Clustering gene expression data}
For grouping set of genes and/or grouping experimental conditions having similar gene expression pattern, clustering algorithms are the most popularly applied method.
Some of the most widely used clustering algorithms are hierarchical, k-means, self organizing maps, fuzzy c-means, and so on \cite{raza14}.
However, FCA has also been used for grouping genes, as an alternative approach to clustering. Choi et al. (2008) \cite{choi08} proposed FCA-based approach
for grouping genes based on their gene expression pattern.  FCA builds a lattice from the gene expression data together with some additional
biological information, where each vertex corresponds to a subset of genes which are clubbed together based on their expression values and
some other functional information. The lattice structures of gene sets are assumed to show biological relationship in the gene expression dataset.
Here, similarities and dissimilarities between different experiments are determined by corresponding lattices. This approach consists of three main
steps: i) building a binary relation, ii) construction of concept lattice, and iii) defining a distance measure and comparing the lattices.
In the first step, the objects are genes, their discretized gene expression attributes and biological attributes.
In the second step, for each experiment a binary relationship is constructed using concept lattice algorithm. Finally, third step calculates distance and compares the lattices.
The work of Choi et al. (2008) \cite{choi08} is an attempt to apply FCA for gene clustering but the distance measure employed was quite
fundamental and it did not properly exploit the properties of the lattice structure. Hence, other possible distance measures such as
spectral distance, maximal common sublattice based distance, etc. can also be investigated \cite{choi08}.
In addition to global lattice comparison, local structure (sublattice) can also be investigated that may assist in identification of particular biological pathways.

Melo and collaborators \cite{melo13} proposed an FCA-based approach combined with association rule and visual analytics to find out overlapping groups of genes in gene expression
and analyzed it in an analytical tool called CUBIST. The workflow of CUBIST involves querying a semantic databases and transforming the result into formal context and then it
is visualized as a concept lattice and associated charts. The CUBIST tool address the challenges of gene expression analysis by filtering and grouping large amount of
 datasets, interactive exploration of data and presents various relevant statistics.

\subsection{Clustering multi-experiment expression data}
Due to availability of high-throughput techniques, presently we have large number of gene expression datasets. Combining datasets taken from
multiple microarray experiments is research question. It has been proved and suggested by many recent studies that the analysis and integration
 of multi-experiment datasets are expected to give more accurate, reliable and robust results. The reason is that integrated datasets would be based
 on large number of gene expression samples and the effects of individual study-specific biases are reduced. For the consensus integration of multi-experiment
  expression data, FCA has been successfully applied by Hristoskova and collaborators \cite{hristoskova14}. They proposed a generic consensus clustering
  which applied FCA for consolidation and analysis of clustering solutions taken from multiple microarray experiments. Initially, the datasets are broken into multiple
   groups of related experiments based on some predefined criteria. In the next step, a consensus clustering technique is deployed to each group that results on
    clustering solution per group. Further, these solutions are pooled together and analyzed by FCA that enables extracting valuable insights from the data and
    generate a gene partition over all the experiments. The FCA-enhanced consensus clustering algorithm proposed by Hristoskova and collaborators \cite{hristoskova14} is depicted
    in Fig. 1. The algorithm is divided into three steps: initialization, clustering and FCA-based analysis. In the initialization step, multi-experiment data are
    divided into $r$ groups of related datasets. Clustering step applies consensus clustering that generates $r$ different solutions. FCA-based analysis step construct
    concept lattice that partitions the genes into a set of disjoint clusters, as shown in Fig. 1. The advantages of FCA-enhanced clustering approach proposed in
     \cite{hristoskova14} are as follows: i) Uses all data that allow each group of related experiments to have a different set of genes, i.e., total set of
     studies genes is not limited to those present in all the datasets, ii) it can be better tuned for each samples by identifying initial number of clusters for
      each group of related experiment, depending upon the number, composition and quality of expression profiles, and iii) the problem with ties is avoided by
       applying FCA to analyze together all partitioned results and find out the final clustering solution representation as the entire experiment collection.

\begin{figure}[t]
\includegraphics[width=12cm]{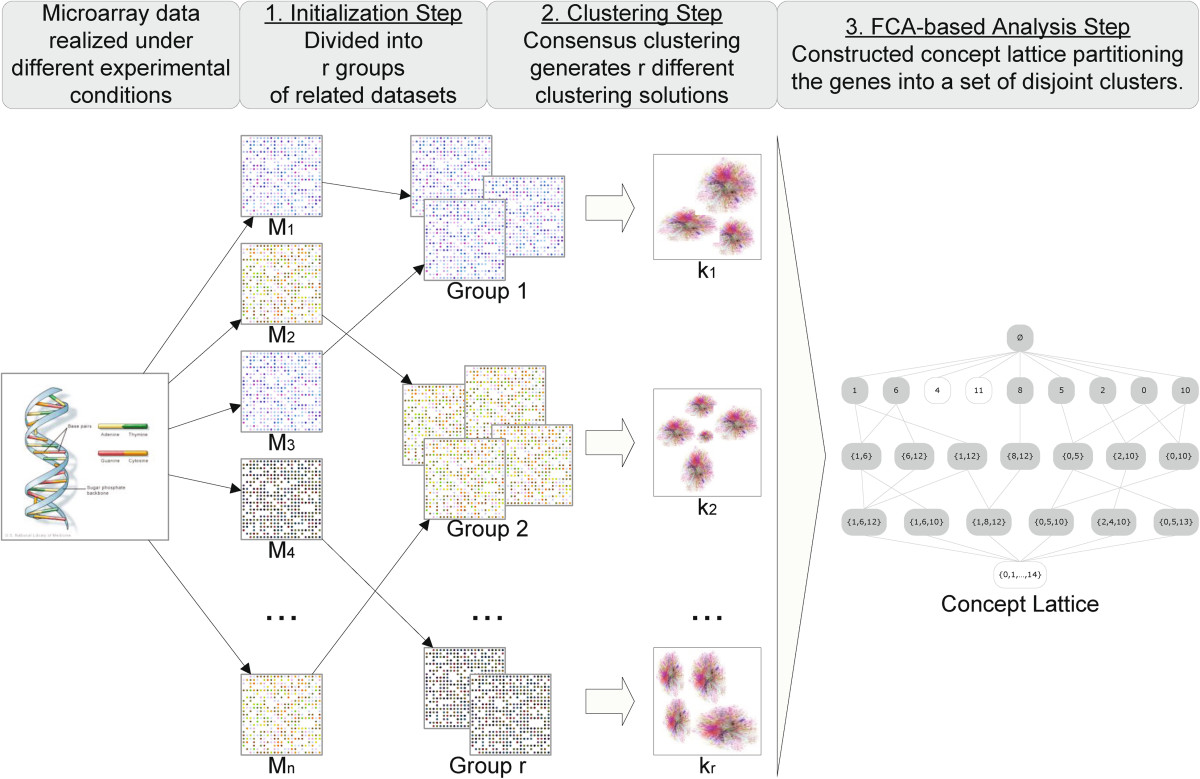}
\caption {Schematic representation of the FCA-enhanced consensus clustering algorithm \cite{hristoskova14}}
\centering
\end{figure}

One another attempt for the application of FCA for knowledge discovery and knowledge integration from gene expression data has been done by
Benabderrahmane (2014) \cite{benabderrahmane14}. Benabderrahmane \cite{benabderrahmane14} introduced a symbolic data mining approach based on
FCA involving bi-clustering of genes, for knowledge discovery and knowledge integration. Firstly, datasets are represented as a formal context
(objects × attributes), where objects are genes and attributes are their expression profiles plus additional information was used such as GO terms that
they annotate, the list of pathways they are involved and their genetic interactions. The algorithm has eight steps, the outline of the algorithm
is depicted in Fig. 2. This algorithm integrates different kinds of datasets such as genes having similar expression profiles and share similar
biological function (GO ontology), knowledge-base of pathways and interactors (KEGG, BioGrid, STRING, etc.)

\begin{figure}[t]
\includegraphics[width=12cm]{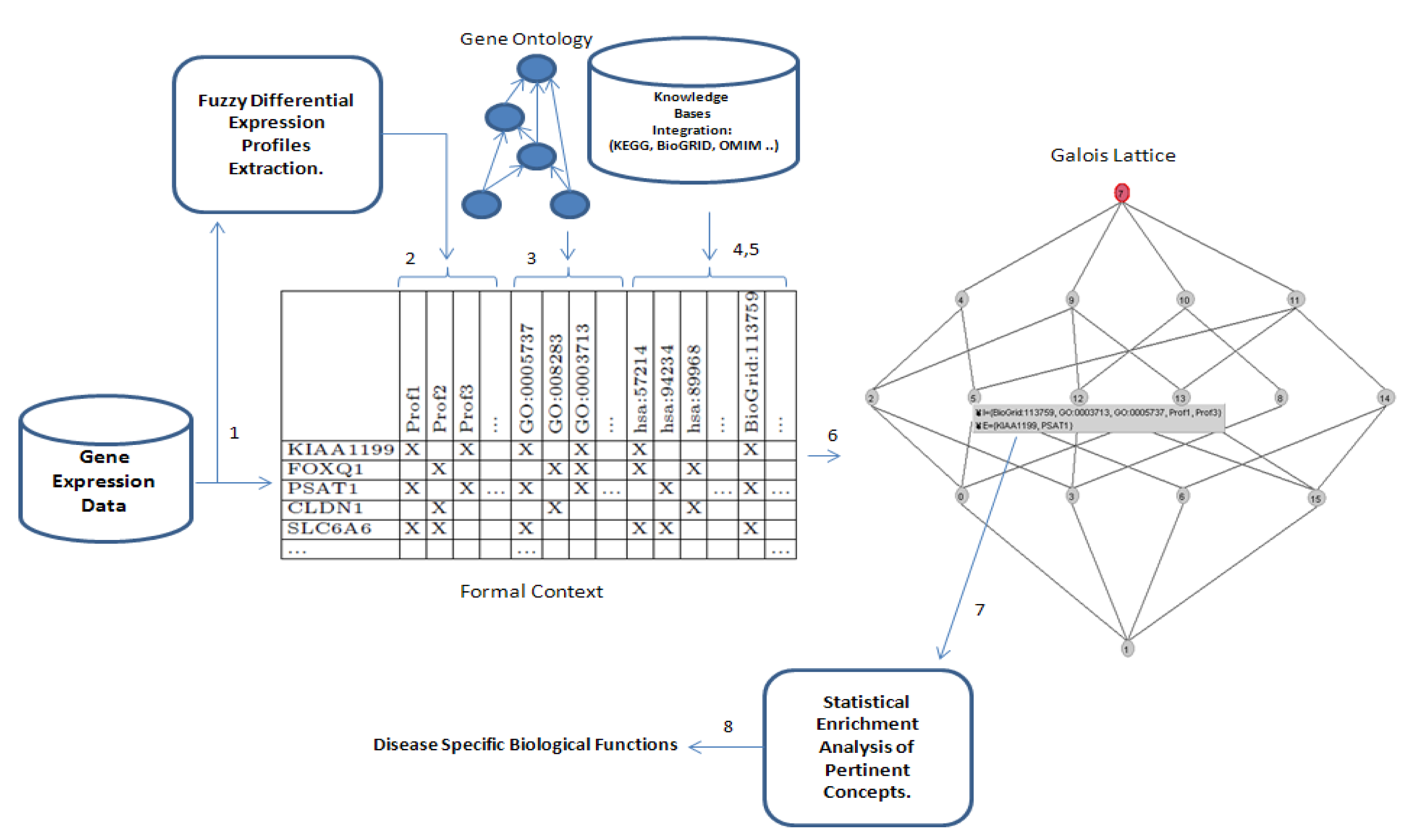}
\caption {An overview of the proposed framework proposed by Benabderrahmane (2014)\cite{benabderrahmane14}}
\centering
\end{figure}

\subsection{Gene expression data comparison}
Finding and understanding the similarities among various diseases is an import research problem in translational bioinformatics. Understanding
disease similarities may help us in refining disease classification, identifying common etiology of comorbidities in genetic studies and
finding analogies between closely related diseases and finally identify common treatments \cite{keller12}.
Bhavnani and collaborators \cite{bhavnani09} applied network analysis approach to find similarities among renal disease using gene expression data.

In addition to many computational techniques, FCA has also been applied for finding disease similarities.
The work of Keller et al., (2012) \cite{keller12} shows the application of FCA for identification of disease similarity.
They identified formal concepts using gene disease associations that indicate hidden relationship among diseases having
same set of associated genes, and gene that are associated with same set of disease. The FCA approach has advantages over network analysis
approach, such as i) FCA allows representation of relationships among several diseases, ii) it provide results in algebraic form allowing
to consider relationship among concepts, and iii) additional gene annotation can be added to refine concepts that assist for the identification
of functional gene relationships within disease groups. FCA has been applied on renal disease dataset that finds unexpected relationships among
disease which are promising but it suffers from few disadvantages. The difficulty with FCA is that many of the formal concepts may not be useful
because only a few formal concepts indicate relationships.

\subsection{Identifying genes of gene regulatory networks}
Gene regulatory networks (GRNs) are the systematic biological networks describing interaction among set of genes in the form of a graph,
where node represents genes and edges defines their regulatory interactions. Understanding the GRNs helps in understanding interactions among genes,
biological and environmental effects and to identify the target genes for drug against the diseases. GRNs have been proved to be a very useful tool used
to describe and explain complex dependencies between key developmental transcription factors (TFs), their target genes and regulators \cite{raza12, raza13}.
 For the better understanding of a gene regulatory network (GRN), it is necessary to know set of genes belonging to it. Identification of these set of
 genes correctly is a challenging task, even for a small subnetworks. In fact, only few genes of a GRN are known and rest of the genes
 are guessed based on experience or informed speculation \cite{gebert08}.  Hence, it is better to rely on experimental data to support these guesses.

Gebert and collaborators \cite{gebert08} presented a new FCA based method to detect unknown members of GRN using time-series gene expression data.
Suppose that $G = \{g_1, g_2, …, g_n\}$ is the set of all genes in an organism and $S \subseteq G$ is set of seed genes.
The goal is to find subset $S' \subseteq G \backslash S$ of genes which interact strongly with GRN defined by set $S$.
Let $R \subseteq G \times G$ be a relation having interactions and $M$ an $n \times l$ matrix that consists of time-series
gene expression profiles having length $l$. If pair $(g_i, g_j) \in R$ then it is known that $g_i$ and $g_j$ interaction to each other.
The FCA-based approach proposed by Gebert et al., (2008) \cite{gebert08} has three main steps described as follows.
First step is preprocessing step that uses the relation $R$ to get an initial list of interesting genes.
If interaction data are not available, this step is skipped and entire gene set $G$ is taken as the initial list of genes.
In the second step, concept lattice is constructed using gene expression data that reduces the number of genes on the initial list.
The last step computes probabilities for the correlation coefficient between genes that result from the second step
and genes of $S$ in order to get list of significant interactions.

\section{Classification and prediction of enzymes, ligand and domain-domain identification}
\label{sec:4}
The classification and study of relations in FCA is focused on the basis of the objects and various types of related attributes
(binary, nominal, ordinal etc.), therefore it is quiet helpful for computational scientist working on Biological data, who may wish to skip the inside details.
With several advantages, including strong mathematical basis, FCA serves in several applications to explore biological data,
enzyme classifications, identification of important protein domains (including protein binding sites) and related drug molecules.
FCA is also reported useful in the integration of Biological activity with chemical spaces. This list is not exhaustive;
FCA has also been used to understand the structural classification of glasses \cite{bartel97} and several other studies.
In this section, we discussed some important application of FCA for the classification of enyzmes, binding site identification and discovery of ligand as drug molecules and so on.

\subsection{Enzyme classification}
Enzymes are proteins which catalyses biological reactions and they are named and classified according to the reaction they catalyse.
For example, hydrolases are those types of enzymes which are involved in the reactions by addition or deletion of water molecules.
Though the sequences of most of the enzymes are available in numerous biological databases, it is tedious task to predict the function of
the enzymes from their respective sequences due to varied activity from small sequence combinations. Considering that the new
enzyme family may emerge, an effort was made for enzyme classifications using FCA which classifies the enzymes
that does not belong to known family \cite{coste14}.
They comment: it is easier to predict the super-families of the proteins as compared to the families of the proteins.
In this study, the labelled and unlabelled enzyme sequences were ‘objects’ whereas ‘attributes’ represent the enzyme blocks.
Enzyme blocks are formed by sequential arrangements of the amino acids, which correspond to specific functions like catalytic site,
lining residues of important pockets or binding sites. In this method of classification, more than half unlabelled sequences were found to be correctly classified.

Another attempt for the classification of protein using FCA has been done by Han and collaborators \cite{han07}. They proposed FCA-based
approach for protein classification that uses protein domain and Gene-Ontology annotation information. Protein domains represent the evolutionary
information forming a protein, while Gene-Ontology describes other properties of proteins that includes structure of protein, molecular interactions, etc.
Han and collaborators \cite{han07} applied tripartite lattice for interpenetrations among protein, domain and GO terms. With the help of
tripartite lattice, they classified protein from domain composite and their corresponding GO term description. They extracted concrete
information using tripartite lattice in the corresponding domain that co-occur in proteins because they are more likely to exhibit
common functions, as annotated in GO terms.

\subsection{Binding site identification}
Protein binding sites (PBS) and ligand binding sites identification are vital to protein- protein and protein-ligand interactions, respectively.
This eventually helps the medical science in identification of better drug or therapy for several important diseases.
There are several ways to identify the binding sites. Most commonly, the protein docking protocol helps in identifying the binding site by
forming complex with one protein to other protein or a ligand (which is a drug in most of the cases).

Bresso et al., (2012) \cite{bresso12} in their report highlight: Majority of the reported methods utilising the structure based prediction methods
for protein-protein interactions consider the attributes, which are physico-chemical properties like hydrophobicity, residue constituents but lack the
representation of properties (e.g. accessible surface of a particular residue) of binding components or spatial relation between two components (residues).
Considering these limitations and knowing the flexibility of FCA, Bresso et al., (2012) \cite{bresso12} utilised available protein 3D structures for characterizing PBS.
In this concept, Inductive Logic Programming (ILP) was linked with FCA, which enabled identifications and discovery of distinct binding pockets of protein-protein interactions.

\subsection{Discovering Ligand from database as a drug molecule}
Using FCA, several attempts have been made to identify suitable ligands from number of chemical database like IUPHAR,
ZINC and many more. The reports suggest that FCA helps in the identification of drug molecules.
Drug molecules can be either agonist (activators) or antagonist (inhibitors).
For a given protein, these drug molecules, would likely act as agonist or antagonist.
The one which do not binds and do not show the changes, are not considered as drug molecules.
In addition to the ADMETox properties, the chemical molecules, which follow the Lipinski's Rule, are considered as suitable drugs.

Actually, when we talk about drugs, a chemical compound has number of physical properties: Hydrogen bond donors, acceptors;
rotatable bonds; topological surface area; molecular weight; XlogP and chemical properties: absorption; digestion; metabolism; excretion;
toxicity. Using these properties as attributes for the object \emph{ligand} which could be possibly a drug molecule, one can identify and differentiate
them from a bulk of chemical molecules in the database using FCA. To take an example, similar attempt was made by
Sugiyama et al., in 2012 \cite{sugiyama12} . They considered the physical features discussed above, including the number of Lipinski’s rule
broken to set as attribute in order to identify the ligands from IUPHAR database. They designed an algorithm, LIFT (LIgand Finding via Formal ConcepT Analysis)
for semi-supervised multi-labelled classification from mixed type data. Results of the algorithm were effective and proved to be
efficient system of classification to identify the ligands from the training data.
Fragment Formal Concept Analysis (FragFCA) introduced by Lounkine et al., (2008) \cite{lounkine08} has the ability to identify the selective hits in
high-throughput screening data sets. In the concept design of FragFCA, combinations of molecular fragments are the 'objects' and their 'attributes'
includes the compound activity and potency information.

The effectiveness of better drug identification can be improved, when the attributes classifying the ligands could be slightly updated,
so as to filter non-peptide molecules from the bulk of drug molecules. It has been identified that the peptide molecules have limited in vivo
efficacy due to pharmacological constraints: solubility, stability and selectivity. Hence, for reliable and safer drug therapy, discovery and
optimisation of non-peptide inhibitors/drugs is necessary \cite{mugumbate13}. Moreover, in a recent in silico identifications of
the drug molecules for Cathepsin L (SmCL1) of the organism, Schistosoma mansoni responsible for the disease ‘schistosomiasis’, it was revealed
that the non-peptide molecules could be better drug molecules as compared to peptide drugs molecules \cite{zafar15}. The list of popularly used software tool based on FCA is shown in Table 1.

So, to conclude, FCA can set an excellent framework to deal with variety of problems. Before application of the concept on to the
biological data minor optimisations and through understanding of the domain is the need in current study for better research.

\begin{table}
\caption{List of FCA based software tools applied in biological domain}
\label{tab:1}       
%
%
\begin{tabular}{p{0.5cm}p{2.2cm}p{4.8cm}p{2.5cm}}
\hline\noalign{\smallskip}
S.No. & Tool Name & Descriptions & References  \\
\noalign{\smallskip}\svhline\noalign{\smallskip}
1. & ConExp  & Java-based FCA analysis software tool used to generate the lattice diagram. & Serhiy (2000) \cite{serhiy00}\\
2. & FcaStone  & Tool for format conversion and command-line lattice generation. & Priss (2008) \cite{priss08}\\
3. & Contextual Role Editor  & FCA tool that work with Eclipse modeling tool. & Mühle \& Wende (2010) \cite{muhle10}\\
4. & FcaBedrock  & A tool for creating context files for Formal Concept Analysis.
    It can convert existing data sets in flat-file CSV or 3-column CSV, to Burmeister (.cxt) or FIMI (.dat) context files. & Andrews \& Orphanides (2010) \cite{andrews10}\\
5. & Lattice Miner  &  FCA tool for the construction, manipulation and visualization of concept lattices. & Lahcen \& Kwuida (2010) ) \cite{lahcen10}\\
6. & CUBIST  & Gene expression analysis tool that combines FCA with association rule and visual analytics.
    It provides filtering and grouping large data sets, its interactive exploration and provides various relevant statistics. & Melo et al., (2013) \cite{melo13}\\
7. & Galicia  & Galois lattice integrative constructor & Galicia \cite{galicia}\\
8. & OpenFCA  & This project comprises of a set of tools for performing FCA activities, including creation of context, visualization of lattice and
    attribute exploration & Borza et al., (2010) \cite{borza10}\\
9. & LIFT  & LIgand Finding via Formal ConcepT Analysis (LIFT) for semi-supervised multi-labelled classification from mixed type data. & Sugiyama et al., (2012) \cite{sugiyama12}\\
10. & FragFCA  & Fragment Formal Concept Analysis (FragFCA) identifies the selective hits in high-throughput screening data sets. & Lounkine et al., (2008) \cite{lounkine08}\\

\noalign{\smallskip}\hline\noalign{\smallskip}
\end{tabular}
\end{table}

\section{Conclusions and Discussions}
\label{sec:5}
Biological data are growing with unprecedented rate. High throughput technologies
 fuelled in the production of high quality biological data.
These data when analyzed properly then one can discover several fruitful knowledge hidden inside biological data.
Formal Concept Analysis (FCA) is a method based on lattice theory for the analysis of binary relational data and has been found to have potential applications
in many areas of bioinformatics and computational biology, beside other applications.
In this chapter, we presented the current status of FCA for the analysis and
knowledge discovery from biological data including gene expression discretization, gene co-expression mining, gene clustering, finding genes in gene regulatory networks, enzyme/protein classifications, binding site classifications, and so on. It also presented a brief list of FCA-based software tools applied in biological domain and covered some challenges faced so far.

%
%
%

\end{document}